\pdfoutput=1

\documentclass[11pt]{article}

\usepackage{EMNLP2022}

\usepackage{times}
\usepackage{latexsym}

\usepackage[T1]{fontenc}

\usepackage[utf8]{inputenc}

\usepackage{microtype}

\usepackage{inconsolata}

\usepackage{subcaption, booktabs}
\usepackage{graphicx}
\usepackage{xcolor}
\usepackage{colortbl}
\usepackage{bm}
\usepackage{amsmath, amssymb}
\usepackage{fdsymbol}
\usepackage{amsfonts}
\usepackage{tikz}
\usetikzlibrary{fit}
\usetikzlibrary{shapes}

\definecolor{yellow}{RGB}{245, 175, 2}
\definecolor{red}{RGB}{229, 50, 56}
\definecolor{green}{RGB}{134, 184, 23}
\definecolor{blue}{RGB}{0, 100, 210}

\title{\textit{Mask More and Mask Later}: Efficient Pre-training of Masked Language Models by Disentangling the [MASK] Token}

\author{Baohao Liao$^{\clubsuit}$, David Thulke$^{\vardiamondsuit}$ \\
  {\bf Sanjika Hewavitharana$^{\spadesuit}$, Hermann Ney$^{\vardiamondsuit}$, Christof Monz$^{\clubsuit}$} \\
  $^{\clubsuit}$University of Amsterdam, $^{\vardiamondsuit}$RWTH Aachen University, $^{\spadesuit}$eBay \\
  \texttt{\href{mailto:b.liao@uva.nl}{b.liao@uva.nl}, \href{mailto:thulke@hltpr.rwth-aachen.de}{thulke@hltpr.rwth-aachen.de}} \\
  \texttt{\href{mailto:shewavitharana@ebay.com}{shewavitharana@ebay.com}, \href{mailto:ney@informatik.rwth-aachen.de}{ney@informatik.rwth-aachen.de}, \href{mailto:c.monz@uva.nl}{c.monz@uva.nl}}
  }

\begin{document}
\maketitle
\begin{abstract}
The pre-training of masked language models (MLMs) consumes massive computation to achieve good results on downstream NLP tasks, resulting in a large carbon footprint. In the vanilla MLM, the virtual tokens, [MASK]s, act as placeholders and gather the contextualized information from unmasked tokens to restore the corrupted information. It raises the question of whether we can append [MASK]s at a later layer, to reduce the sequence length for earlier layers and make the pre-training more efficient. We show: (1) [MASK]s can indeed be appended at a later layer, being disentangled from the word embedding; (2) The gathering of contextualized information from unmasked tokens can be conducted with a few layers. By further increasing the masking rate from 15\% to 50\%, we can pre-train RoBERTa-base and RoBERTa-large from scratch with only 78\% and 68\% of the original computational budget without any degradation on the GLUE benchmark. When pre-training with the original budget, our method outperforms RoBERTa for 6 out of 8 GLUE tasks, on average by 0.4\%. \footnote{Code at \url{https://github.com/BaohaoLiao/3ml}}
\end{abstract}

\section{Introduction}
\label{sec: introduction}
Large-scale pre-trained MLMs, like BERT \cite{DBLP:conf/naacl/DevlinCLT19} and its variants \cite{DBLP:journals/corr/abs-1907-11692, DBLP:conf/iclr/LanCGGSS20, DBLP:conf/iclr/ClarkLLM20, DBLP:conf/icml/SongTQLL19, DBLP:conf/acl/LewisLGGMLSZ20}, have achieved great success in various NLP tasks, such as machine translation \cite{DBLP:journals/tacl/LiuGGLEGLZ20, DBLP:conf/iclr/ZhuXWHQZLL20}, general language understanding \cite{DBLP:conf/iclr/WangSMHLB19, DBLP:conf/nips/WangPNSMHLB19}, question answering \cite{DBLP:conf/emnlp/RajpurkarZLL16}, summarization \cite{DBLP:journals/corr/abs-1903-10318} and claim verification \cite{DBLP:conf/ecir/SoleimaniMW20}. To make the pre-trained model generalize well to a wide range of tasks, MLMs tend to have a large number of parameters, even in the billion scale \cite{DBLP:journals/corr/abs-1909-08053}, and are trained with plenty of data. This is prohibitively expensive and generates significant amounts of CO$_2$ emissions \cite{DBLP:conf/acl/StrubellGM19, DBLP:journals/corr/abs-2104-10350}. How to make the pre-training of MLMs more efficient while retaining their superior performance is a critical research question.

Various attempts on efficient pre-training have obtained effective results. \citet{hou-etal-2022-token} and \citet{DBLP:conf/iclr/0001XLKHL21} applied a prior knowledge extracted from the MLM itself to make the prediction more inclined to rare tokens. \citet{DBLP:journals/corr/abs-1909-08053} and \citet{DBLP:conf/iclr/YouLRHKBSDKH20} made use of mixed-precision and distributed training to speed up the pre-training. Data-efficient pre-training objectives \cite{DBLP:conf/iclr/ClarkLLM20, DBLP:conf/iclr/LanCGGSS20} and progressively stacking technique \cite{DBLP:conf/icml/GongHLQWL19} also work quite well. Orthogonal to these directions, we dive deeply into the information flows of the vanilla MLM, trying to split different types of information flows into multiple stages and making the model spend more computation on the complex flow.

\begin{figure}[t]
  \centering
\begin{tikzpicture}
    \node[minimum height=2em] (A) at (0,0) {\Large $x_1$};
    \node[right, minimum height=2em] (B) at (A.east) {\Large $m_2$};
    \node[right, minimum height=2em] (C) at (B.east) {\Large $m_3$};
    \node[right, minimum height=2em] (D) at (C.east) {\Large $x_4$};
    \node[right, minimum height=2em] (E) at (D.east) {\Large $x_5$};

    \path [stealth-, bend left=90, thick, dashed, yellow] (A.north) edge (B.north);  
    \path [stealth-, bend left=90, thick, dashed, yellow] (A.north) edge (C.north);  
    \path [stealth-, bend right=90, thick, dashed, yellow] (D.north) edge (B.north);  
    \path [stealth-, bend right=90, thick, dashed, yellow] (D.north) edge (C.north);  
    \path [stealth-, bend right=90, thick, dashed, yellow] (E.north) edge (B.north);  
    \path [stealth-, bend right=90, thick, dashed, yellow] (E.north) edge (C.north);
    \path [-stealth, bend right=90, thick,blue] (A.south) edge (B.south);  
    \path [-stealth, bend right=90, thick,blue] (A.south) edge (C.south);  
    \path [-stealth, bend left=90, thick,blue] (D.south) edge (B.south);  
    \path [-stealth, bend left=90, thick,blue] (D.south) edge (C.south);  
    \path [-stealth, bend left=90, thick,blue] (E.south) edge (B.south);  
    \path [-stealth, bend left=90, thick,blue] (E.south) edge (C.south);    
\end{tikzpicture}
\begin{tikzpicture}
    
    \node (A) at (0,0) {\Large $x_1$};
    \node[below left, yshift=-8mm, xshift=-1mm] (B) at (A.south west) {\Large $x_4$};
    \node[below right, yshift=-8mm, xshift=1mm] (C) at (A.south east) {\Large $x_5$};
    \node (D) at (3.5,0) {\Large $m_2$};
    \node[below, yshift=-8mm] (E) at (D.south) {\Large $m_3$};
    
    \node[fill=blue!33,fit=(A) (B) (C),ellipse, inner sep=-1mm, yshift=-1mm] (U) {};
    \node[fit=(D) (E), inner sep=-1mm, yshift=-1mm, minimum width =2em] (M) {};
    \node[fill=yellow!20,fit=(D) (E),ellipse, inner sep=-1mm, minimum width =2em] (M_) {};
    
    \node[minimum height=2em] (A_l) at (A) {\Large $x_1$};
    \node[minimum height=2em] (B_l) at (B) {\Large $x_4$};
    \node[minimum height=2em] (C_l) at (C) {\Large $x_5$};
    \node[minimum height=2em] (D_l) at (D) {\Large $m_2$};
    \node[minimum height=2em] (E_l) at (E) {\Large $m_3$};
    
    \path [stealth-stealth, thick] (A.south west) edge (B.north);
    \path [stealth-stealth, thick] (A.south east) edge (C.north);
    \path [stealth-stealth, thick] (B.east) edge (C.west);  
    \path [stealth-stealth, thick, dashed] (D.south) edge (E.north);  
    
    \path [-stealth, thick, blue, shorten >=2mm, shorten <=12.5mm] (A.east) edge (D.west);  
    \path [stealth-, thick, dashed, yellow, shorten >=2mm, shorten <=4mm] (C.east) edge (E.west);  
    
\end{tikzpicture}
\begin{tikzpicture}
 \path ([yshift=-1em]current bounding box.south)
 node[matrix,cells={nodes={font=\sffamily,anchor=west}}, row sep=-2mm]{
  \draw[-stealth,color=black,thick](0,0) -- ++ (0.3,0); & \node{\small{/}}; & \draw[-stealth,color=blue,thick](0,0) -- ++ (0.3,0); & \node{\tiny{token \& positional information}};\\
  \draw[-stealth,color=black,dashed,thick](0,0) -- ++ (0.3,0); & \node{\small{/}}; & \draw[-stealth,color=yellow,dashed,thick](0,0) -- ++ (0.3,0); & \node{\tiny{positional information}};\\
 };

\end{tikzpicture}
    \caption{\textbf{Information flows of vanilla MLM}. A sentence, $\{x_1,  x_2, x_3, x_4, x_5\}$, is corrupted by replacing $x_2$ and $x_3$ with a virtual token $m$ indexed with the corresponding positions. {\textcolor{white}{\footnotemark}} }
    \label{fig: information flow}
\end{figure}
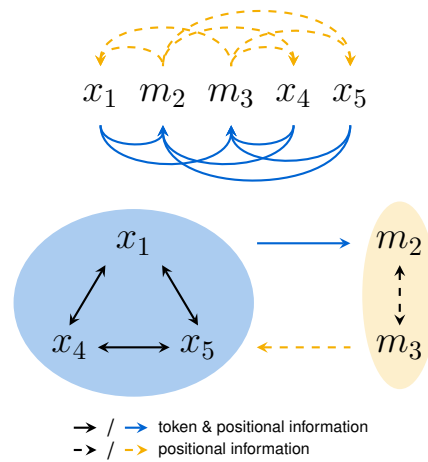
\footnotetext{Except where explicitly mentioned, we ignore randomly replaced and unchanged tokens for simplicity.}

The information transferred among tokens can be split into: \textit{position information} and non-positional information (termed as \textit{token information} in this paper). As shown in Figure \ref{fig: information flow}, the information flows for a corrupted sentence during training consist of: (1) Position and token information among unmasked tokens; (2) Position and token information from unmasked tokens to [MASK]s; (3) Position information among [MASK]s; (4) Position information from [MASK]s to unmasked tokens. These information flows happen in each Transformer block \cite{DBLP:conf/nips/VaswaniSPUJGKP17}, more specifically in the self-attention module. In addition, the 4$^{th}$ flow brings no additional information given the 1$^{st}$ one, since the positions information from [MASK]s can be inferred implicitly given the positions of the unmasked tokens. We ignore the 4$^{th}$ flow for the following discussion. 

Intuitively, the amount of information transferred in each flow is not at the same level. The 3$^{rd}$ flow contains the least information. We make an assumption about the first two flows. \\

\noindent \textbf{The Information Flow Assumption.} \textit{ Position and token information among unmasked tokens (1$^{st}$ flow) are more difficult to learn than the transfer of this knowledge to [MASK]s (2$^{nd}$ flow).}
\\

\noindent This assumption is empirically proven later (\S \ref{sec:mask_later}). Since the difficulty of information transfer varies among different flows, it makes sense to divide the flows into multiple stages, forcing the model to spend more computation on the more complex one. 

We propose a two-stage learning method. For the early layers of an MLM, we detach [MASK]s and only input the embedding of unmasked tokens. So the model firstly focuses on the most complex (1$^{st}$) information flow. At an intermediate layer, we append the embedding of [MASK]s with their corresponding position information back to the sequence. Then the remaining layers of the MLM fuse all information. In this way, the sequence length for the early layers becomes shorter due to excluding [MASK]s. We further reduce the sequence length by increasing the masking rate for higher efficiency. We call our method \textit{mask more and mask later (3ML)}, since [MASK]s are appended later and we have a larger masking rate. 

In this work, we introduce two models designed for 3ML (\S \ref{sec: method}), empirically show two prerequisites of efficiency for 3ML hold (\S \ref{sec: prerequisites}), conduct extensive experiments to select an optimal setting for high performance and efficiency (\S \ref{sec: efficient setting}), and finally compare 3ML's results to strong baselines (\S \ref{sec: compare with previous work}).

Our main contributions are summarized as: (1) We introduce a simple, intuitive but effective method for the efficient pre-training of MLM; (2) We prove two prerequisites that are important for 3ML; (3) On the GLUE benchmark, 3ML achieves the same performance as RoBERTa-base and RoBERTa-large \cite{DBLP:journals/corr/abs-1907-11692} with only 78\% and 68\% of the original computation budget, and outperform them with the same budget.

\begin{figure*}[h]
  \centering
    \includegraphics[width=1.0\textwidth]{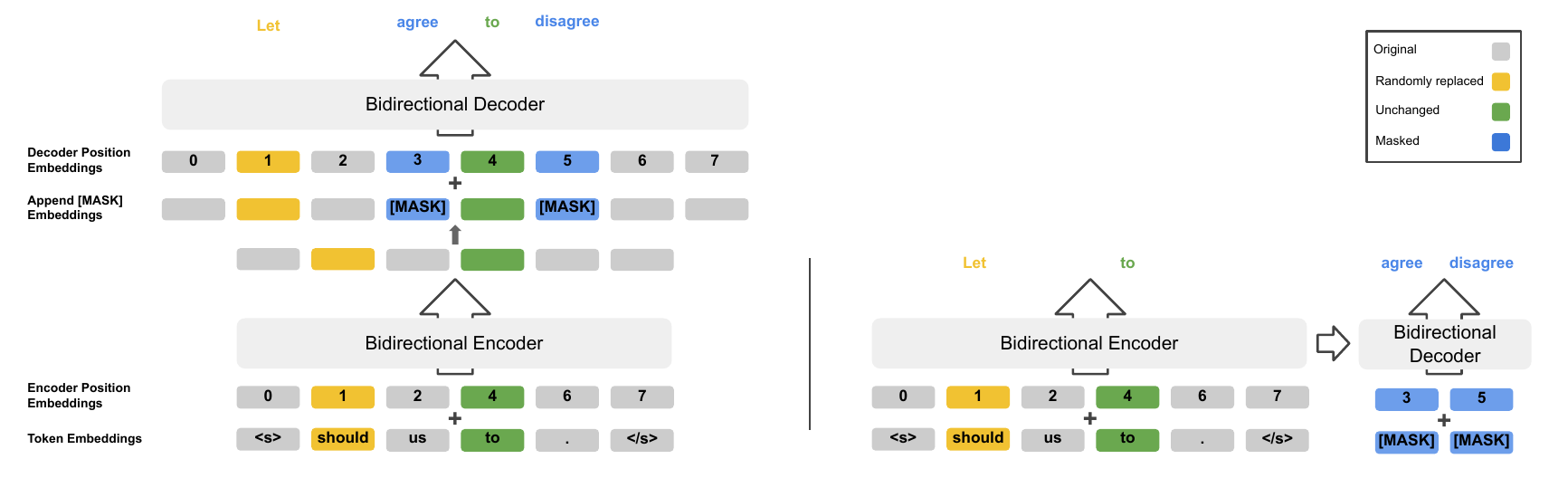}
    \caption{\textbf{Overview of 3ML architectures}. A sentence,``<s> Let us agree to disagree. </s>'', is corrupted to ``<s> should us [MASK] to [MASK]. </s>''.  Left: 3ML with only self-attention layers (3ML\textsubscript{self}). Right: 3ML with a decoder consisting of both self- and cross-attention layers (3ML\textsubscript{cross}). We achieve efficient pre-training by discarding [MASK]s for the encoder and applying a small decoder for the whole sequence length. For fine-tuning on downstream tasks, the decoder is removed.}
    \label{fig: 3ml architecture}
\end{figure*}

\section{Model}
\label{sec: method}
In this section, we first discuss the information flows in a vanilla MLM, i.e. BERT, then introduce two architectures designed for 3ML (Figure \ref{fig: 3ml architecture}). 

\subsection{Vanilla Masked Language Model}
Masked language models reconstruct a sequence with corrupted information. Given a sequence of tokens $\bm{x} = \{x_t\}_{t=1}^T$ with $t$ denoting the token's position, the corrupted version $\bm{\hat{x}}$ is generated by randomly setting a portion of $\bm{x}$ to a special symbol [MASK]. MLM is trained to learn the distribution $p(\bm{x} | \bm{\hat{x}})$ with a loss function:
\begin{equation}
  \mathcal{L} = \mathop{\mathbb{E}_{\bm{x} \sim \mathcal{D}}[- \sum_{t=1}^{T} \delta_{x_t \neq \hat{x}_t} \log p_{\theta}(x_t | \bm{\hat{x}})]}
\end{equation}
where $\delta_{x_t \neq \hat{x}_t}$ is a Kronecker Delta function:
\begin{equation}
  \delta_{x_t \neq \hat{x}_t} =
  \begin{cases}
    1 & \text{$x_t \neq \hat{x}_t$} \\
    0 & \text{$x_t = \hat{x}_t$}
  \end{cases} \nonumber
\end{equation}

As shown in Figure \ref{fig: information flow}, position and token information are transferred among different tokens in the model. We can cluster the flows into four types: (1) From unmasked tokens to unmasked tokens (within the blue area): MLM transfers position and token information among unmasked tokens, generating uncorrupted contextualized information; (2) From unmasked tokens to [MASK]s (from the blue area to the yellow area): MLM transfers the uncorrupted contextualized information to [MASK]s; (3) From [MASK]s to [MASK]s (within the yellow area): MLM transfers the position information among [MASK]s. Since all masked tokens have the same token embedding, there is no transfer of token information; (4) From [MASK]s to unmasked tokens (from the yellow area to the blue area): MLM transfers position information from [MASK]s to unmasked tokens.


\subsection{Mask More and Mask Later}
\label{sec: 3ML}
Since different flows of the vanilla MLM contain different amounts of information (\S \ref{sec: introduction}), we propose a two-stage training method where more computation is allocated to the most complex flow, the one among unmasked tokens. At a later stage, we fuse all flows together as vanilla MLM. In this way, we aim to improve the efficiency by reducing the sequence length for the first stage by discarding [MASK]s. Even though at the later stage we still need to fuse all information flows together, back to the original sequence length, we only need to apply a few layers for that, since the 1$^{st}$ flow is already learned quite well during the first stage. Combining a large masking rate and a small number of layers for the second stage together, we can achieve an efficient pre-training. In short, two prerequisites contribute to our efficient pre-training: we can mask more and mask later (test in \S \ref{sec: prerequisites}). 

We design two architectures, \textit{3ML\textsubscript{self}} and \textit{3ML\textsubscript{cross}}, that only differ from each other on the decoder. As shown in Figure \ref{fig: 3ml architecture}, we first input the unmasked tokens to both models. At an intermediate layer, we input the token embedding of [MASK] and fuse all information flows together. With our method, the token embedding of [MASK] is disentangled from the original word embedding space and located in a latent space. 

\textbf{3ML\textsubscript{self}} This architecture is inspired by a computer vision method \cite{DBLP:journals/corr/abs-2111-06377} designed for two-stage learning. 3ML\textsubscript{self} has an encoder and a decoder, with a self-attention Transformer block as the base layer for both. A prediction layer for masked tokens is located at the end of the decoder. Only unmasked tokens are fed into the encoder. So the encoder only transfers information among unmasked tokens ($1^{st}$ flow). After the encoder, we append the token embedding of [MASK] back to the sequence with a new positional embedding and input to the decoder. Since the input sequence to the decoder consists of the representations of unmasked tokens and the token embedding of [MASK]s, the decoder fuse all information together. In addition, the token embedding of [MASK] is in the same space as the latent representations of unmasked tokens. 

\textbf{3ML\textsubscript{cross}} Both 3ML\textsubscript{self} and 3ML\textsubscript{cross} share the same encoder that only receives unmasked tokens as input. In contrast to 3ML\textsubscript{self}, we use a Transformer block with both self- \& cross-attention as the base layer for the 3ML\textsubscript{cross} decoder. Two sequences are given as input to the decoder, the latent representations of unmasked tokens and a sequence of [MASK] tokens and their position embeddings. The information flow from unmasked tokens to [MASK]s ($2^{nd}$ flow) are conducted by the cross-attention module. The information among [MASK]s ($3^{rd}$ flow) is transferred by the self-attention module. But there is no further information transfer among unmasked tokens ($1^{st}$ flow) in the decoder, different from 3ML\textsubscript{self}. We use the same prediction layer for the randomly replaced, unchanged and masked tokens. Noticeably, we predict the randomly replaced and unchanged tokens after the encoder.

For both 3ML\textsubscript{self} and 3ML\textsubscript{cross}, if the hidden dimensions of the encoder and decoder are not identical, one extra fully-connected layer is added in between for projection. The prediction layer of vanilla MLM contains two fully-connected layers. The last one shares the weight from the embedding layer that has the same hidden dimension as the encoder. It projects the hidden dimension to the vocabulary size for prediction. If the hidden dimensions of the encoder and decoder are different, the first prediction layer projects the hidden dimension of the decoder back to the encoder dimension, so we can still share the weight from the embedding layer. More details are in Appendix \ref{sec: FLOPs}.

\textbf{Fine-tuning and Inference} After pre-training, the decoder of both architectures is removed. We only fine-tune the encoder on downstream tasks. We implement fine-tuning in this way because we want to: (1) speed up the inference; (2) keep our architecture for downstream tasks the same as standard MLMs and make it convenient for various applications without modifying their frameworks. However, for some tasks that require the [MASK] embedding, like the mask-infilling task, it might be beneficial to keep the decoder, since the token embedding of [MASK] doesn't locate in the same space as other tokens.

\begin{figure*}[t]
  \centering
    \includegraphics[width=1.0\textwidth]{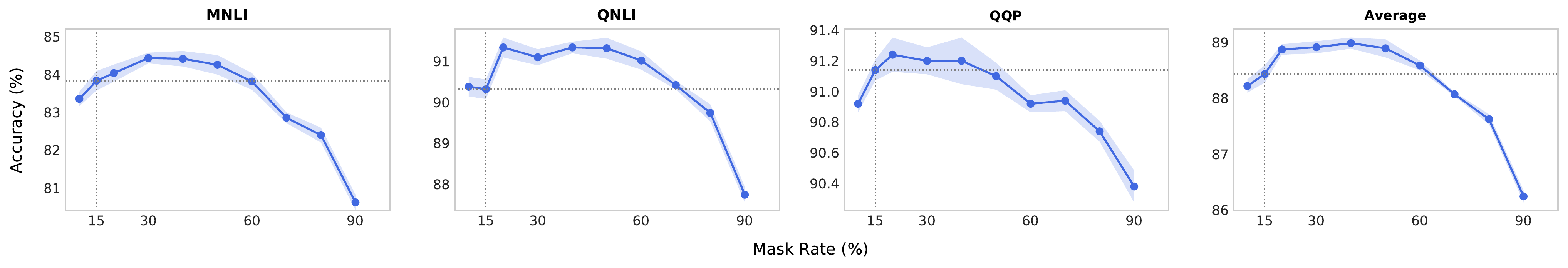}
    \caption{\textbf{Masking rate}. A middle-level masking rate ($40\%$) works best on average.}
    \label{fig: 24hbert mask more}
\end{figure*}

\section{Experimental Setup}
\subsection{Tasks} 
We evaluate our pre-trained models on the General Language Understanding Evaluation (GLUE) benchmark \cite{DBLP:conf/iclr/WangSMHLB19}, which consists of 2 single-sentence classification tasks: CoLA \cite{DBLP:journals/tacl/WarstadtSB19} and SST \cite{DBLP:conf/emnlp/SocherPWCMNP13}, 3 similarity and paraphrase tasks: MRPC \cite{DBLP:conf/acl-iwp/DolanB05}, QQP\footnote{\url{https://www.quora.com/profile/Ricky-Riche-2/First-Quora-Dataset-Release-Question-Pairs}}, and STS \cite{DBLP:journals/corr/abs-1708-00055}, and 4 natural language inference tasks: MNLI \cite{DBLP:conf/naacl/WilliamsNB18}, QNLI \cite{DBLP:conf/emnlp/RajpurkarZLL16}, RTE \cite{DBLP:conf/mlcw/DaganGM05, haim2006second, DBLP:conf/acl/GiampiccoloMDD07, DBLP:conf/tac/BentivogliMDDG09}, and WNLI \cite{DBLP:conf/kr/LevesqueDM12}. Like the original BERT paper \cite{DBLP:conf/naacl/DevlinCLT19}, we exclude WNLI, as the standard fine-tuning approach couldn't even beat the majority classifier.

We report accuracy for SST-2, MNLI, QNLI, and RTE, both F1 score and accuracy for MRPC and QQP, Matthew's correlation for
CoLA, both Pearson and Spearman correlation for STS. By default, we use the same calculation as the GLUE leaderboard, i.e. the average of MNLI-m and MNLI-mm for MNLI, the average of F1 and accuracy for MRPC and QQP, and the average of Pearson and Spearman correlation for STS. We finally report the macro average of all tasks. For ablation experiments, we only evaluate the models on MNLI, QNLI, and QQP and report their accuracy, since these three tasks have the largest amount of training and validation sets, resulting in more stable fine-tuning scores than the others. In addition, the weighted average accuracy for MNLI from MNLI-m and MNLI-mm is shown rather than the macro average for ablation studies\footnote{Both macro average and weighted average scores are comparable since MNLI-m and MNLI-mm contain a similar number of validation samples (9816 and 9833).}. 

\subsection{Baselines} 
We compare 3MLs to the following baselines:

\begin{itemize}
    \item \textbf{Google BERT} The results of Google BERT \cite{DBLP:conf/naacl/DevlinCLT19} on the GLUE development set are not shown in the original paper, we borrow the BERT-base's and BERT-large's results from \citet{xu-etal-2020-bert} and \citet{clark-etal-2019-bam}, respectively.
    
    \item \textbf{Our 24hBERT} 24hBERT \cite{izsak-etal-2021-train} achieves comparable performance to the original BERT \cite{DBLP:conf/naacl/DevlinCLT19} with an academically friendly computation budget (24 hours with 8 Nvidia Titan-V GPUs). Like \citet{DBLP:journals/corr/abs-2202-08005}, we re-implement it by adopting RoBERTa's BPE tokenizer \cite{DBLP:conf/acl/SennrichHB16a, DBLP:journals/corr/abs-1907-11692} for better performance. This is the main baseline for our ablation studies.
    
    \item \textbf{Our RoBERTa} RoBERTa-base's GLUE results (trained on BooksCorpus and English Wikipedia) are not fully shown in the original paper \cite{DBLP:journals/corr/abs-1907-11692}. We re-implement it with its original  hyperparameters. 
    
    \item \textbf{ELECTRA} A discriminatively pretrained language model from \citet{DBLP:conf/iclr/ClarkLLM20}.
\end{itemize}

We don't include the encoder-decoder architectures (like BART \cite{DBLP:conf/acl/LewisLGGMLSZ20} and MASS \cite{DBLP:conf/icml/SongTQLL19}) here because RoBERTa outperforms them \cite{DBLP:conf/acl/LewisLGGMLSZ20} on GLUE tasks.

\subsection{Implementation.}
The re-implementation of baselines and our pre-training methods are conducted on fairseq \cite{DBLP:conf/naacl/OttEBFGNGA19}. All results are from the models pre-trained on BooksCorpus and English Wikipedia that are tokenized by RoBERTa's BPE tokenizer with a vocabulary size of 50K\footnote{This is the main reason for the different model parameters between 3ML and baselines. BERT and ELECTRA use a vocabulary with 30K tokens}. In addition, all results with a single number are the median of five trials. 

\textbf{Training Recipes} We have two pre-training recipes: an efficient recipe for a sequence length of 128 and a longer recipe for a sequence length of 512. The efficient pre-training recipe from 24hBERT \cite{izsak-etal-2021-train} is mainly used for ablation studies. It is a computation-friendly recipe that takes 9 hours with 16 Nvidia Tesla V100 GPUs. The longer pre-training recipe from RoBERTa \cite{DBLP:journals/corr/abs-1907-11692} takes about 36 hours with 32 Nvidia Tesla V100 GPUs. With our efficient method, we can reduce the training time proportionally to the reduced computation (FLOPs). More hyper-parameter details for these two recipes are shown in Table \ref{tab: pretraining hyperparameters} (see Appendix \ref{sec: pretraining hyperparameters}). And the calculation details of training FLOPs are in Appendix \ref{sec: FLOPs}.

The default masking strategy for 3ML stays the same as BERT.
That is for all corrupted tokens, $80\%$ of them are replaced by [MASK], $10\%$ are replaced by random tokens from the vocabulary and $10\%$ stay unchanged. We borrow the same fine-tuning hyperparameters from 24hBERT for all 3MLs (Table \ref{tab: finetuning hyperparameters} in Appendix \ref{sec: finetuning hyperparameters}).

\textbf{Architectures} The encoder of our large or base model shares the same settings as Google BERT. By default, we use a two-layer decoder with half of the hidden dimension of the encoder. 3ML\textsubscript{self} uses Transformer \cite{DBLP:conf/nips/VaswaniSPUJGKP17} encoder layers for its decoder, while 3ML\textsubscript{cross} uses Transformer decoder layers with both self-attention and cross-attention layers without causal masking for its decoder. Since the hidden dimensions of the encoder and decoder are not the same, there is a linear layer in between to project the output from the encoder to the dimension of the decoder. 3MLs have two untied learnable positional embeddings for the encoder and decoder.

Like 24hBERT, we implement pre-layer normalization (pre-LN) \cite{DBLP:journals/corr/abs-1909-08053} for 3ML-large. It makes the pre-training more stable and achieves better performance with a large learning rate. For 3ML\textsubscript{self}-base, post-LN is slightly better than pre-LN. We still use pre-LN for 3ML\textsubscript{cross}-base for stable pre-training. Post-LN doesn't work for 3ML\textsubscript{cross}. We leave the investigation of this problem to future work. For fine-tuning GLUE tasks, the 3ML decoder is removed. So its inference time on downstream tasks stays the same as the original BERT. More details of the 3ML architectures are in Table \ref{tab: pretraining hyperparameters} (see Appendix \ref{sec: pretraining hyperparameters}).

\begin{figure}[t]
  \centering
    \includegraphics[width=0.4\textwidth]{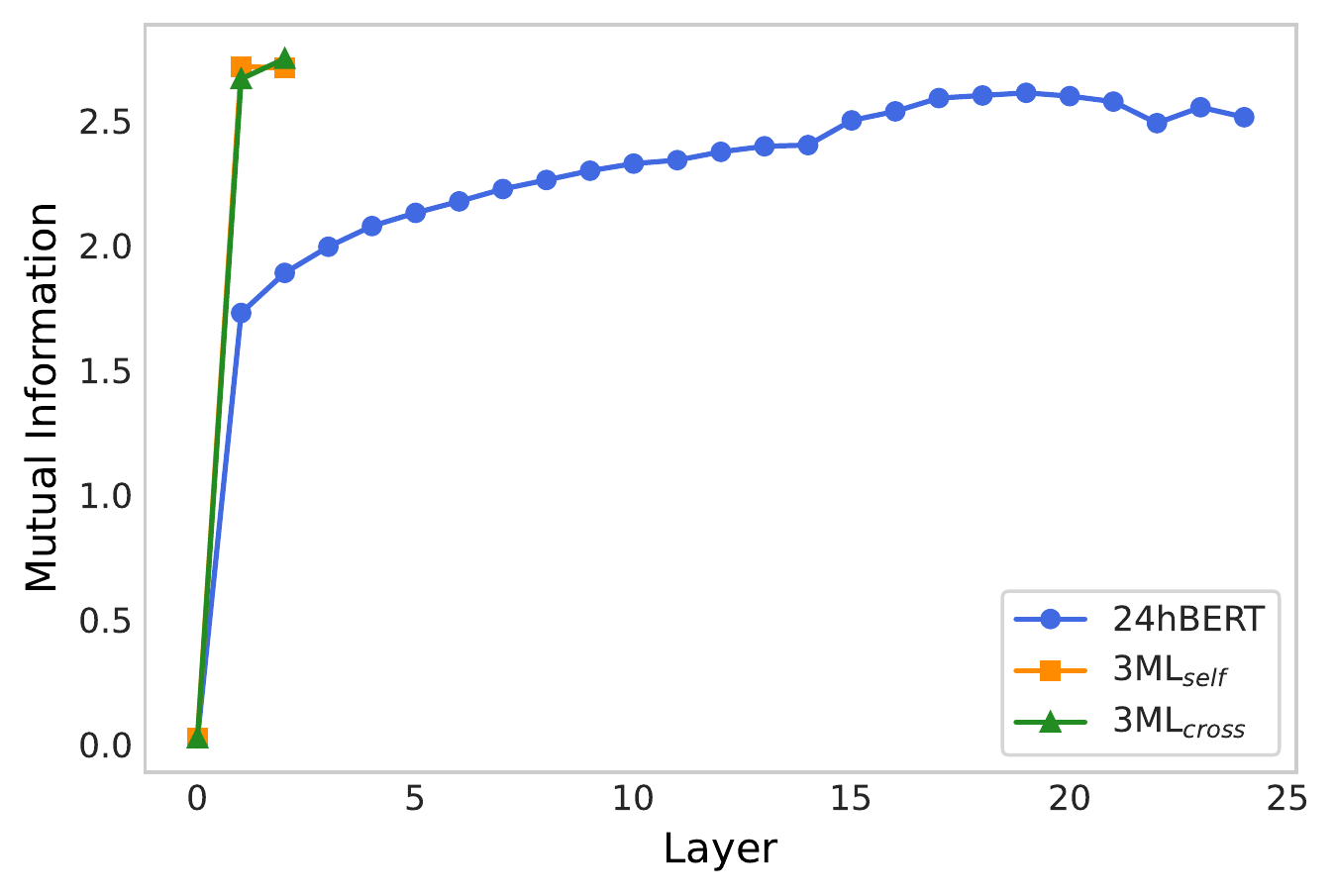}
    \caption{\textbf{Mutual Information} between hidden representations of [MASK] tokens per layer and the original tokens. Layer 0 corresponds to the token embeddings. All models are pre-trained with a masking rate of $40\%$.}
    \label{fig: mutual information}
\end{figure}

\section{Two Prerequisites for Efficiency}
\label{sec: prerequisites}
In this section, we show that the two prerequisites, masking more and masking later, for efficiency hold for 3ML, which is also an empirical test of our information flow assumption.

\subsection{Mask More}
\label{sec: mask more}
As shown in Figure \ref{fig: 3ml architecture}, if masking more is possible, the sequence length of the input to 3ML's encoder becomes shorter. Since most trainable parameters are located in the encoder (the 3ML\textsubscript{self}-large encoder contains 98\% of the parameters, excluding the embedding layer and the prediction layer) and the computation time of a self-attention module scales quadratically with the sequence length, this significantly reduces the computation time. 

However, the motivation of our work is to maintain MLM's performance with higher efficiency. We don't want to lose performance significantly. Therefore, we conduct an experiment on a vanilla MLM, 24hBERT, to check whether masking more is possible.

As shown in Figure \ref{fig: 24hbert mask more}, the default masking rate (15\%) from BERT is not optimal. With an increasing masking rate, the performance of all three tasks increases first and then decreases. The optimal masking rate is 30\% for MNLI, 20\% for QNLI, and 20\% for QQP. The performance of a masking rate in (15, 45)\% is consistently better than the one of 15\%. On average, a masking rate of 40\% works the best. \citet{DBLP:journals/corr/abs-2202-08005} also obtained a similar result. In short, masking more is not only possible but also offers higher performance.

\subsection{Mask Later}
\label{sec:mask_later}
\begin{table}
  \centering
    \begin{tabular}{c|ccc}
    \toprule
    \textbf{Model} & 24hBERT &  3ML\textsubscript{self} & 3ML\textsubscript{cross} \\
    \textbf{PPL} & 7.8 & 9.5 & 9.3 \\
    \bottomrule
\end{tabular}
\caption{\textbf{MLM perplexity} of 24hBERT and 3MLs on the validation set. All models are pre-trained with a masking rate of $40\%$. \label{perplexity}}
\end{table}

Masking later and masking more are complementary to achieve higher efficiency. For vanilla MLM, masking more offers better performance. But it doesn't guarantee that we can disentangle [MASK]s from the word embedding and append them at an intermediate layer. Instead of directly showing the performance of 3ML, we try to answer the following two questions for checking the possibility of masking later: (1) How good are 3MLs at masked language modeling compared to BERT? (2) How fast do the models recover the identity of the [MASK] tokens?

To answer the first question, we compare the MLM perplexity for [MASK]s between 24hBERT and 3MLs on the validation set. The results in Table \ref{perplexity} show:
While 3ML\textsubscript{self} and 3ML\textsubscript{cross} have comparable perplexities, the perplexity of 24hBERT is significantly lower, indicating that 24hBERT performs better at the MLM task. However, this does not necessarily correlate with better performance on downstream tasks due to the mismatch between pre-training and fine-tuning (further discussion in \S \ref{sec: masking strategy}). One could even argue that the increasing difficulty of the pre-training task forces 3MLs to learn better hidden representations of the unmasked tokens.

To address the second question, we measure the mutual information between the hidden representations at each layer at the masked positions and the original tokens at these positions. We follow the strategy proposed by \citet{voita-etal-2019-bottom}, and take hidden representations at masked positions corresponding to the 1000 most frequent tokens. For each layer, we gather 5M hidden representations and cluster them using mini-batch k-means into 10,000 clusters. For 24hBERT, we do this for each of the 24 encoder layers. For the 3ML models, the masked tokens are only fed in the decoder, values are only calculated for two decoder layers.

The results for 24hBERT and 3MLs are shown in Figure \ref{fig: mutual information}. For vanilla MLM, the largest amount of information on the identity of masked tokens is restored after the first few layers. The information is further gradually restored over the remaining layers. The results for 3ML\textsubscript{self} and 3ML\textsubscript{cross} are similar. After the first decoder layer, the information on the masked token identity is already restored to a similar level as in the last layer of 24hBERT. A possible explanation is that the decoder of 3MLs can already access the high-level representation from the encoder that facilitates the reconstruction. 

The results from Figure \ref{fig: mutual information} also empirically test our information flow assumption. The $1^{st}$ flow contains the most significant amount of information. We can achieve higher efficiency by specifically allocating more computation to this flow and spending less computation on the others, rather than focusing on all flows at once like vanilla MLM.

\definecolor{Gray}{gray}{0.85}
\begin{table*}[h]
  \scriptsize
    \begin{subtable}{0.49\textwidth}
        \centering
        \begin{tabular}{c|ccc|ccc}
        \toprule
         & \multicolumn{3}{c|}{\textbf{3ML\textsubscript{self}}} & \multicolumn{3}{c}{\textbf{3ML\textsubscript{cross}}} \\
        \textbf{\#layer} & \textbf{MNLI} & \textbf{QNLI} & \textbf{QQP} & \textbf{MNLI} & \textbf{QNLI} & \textbf{QQP} \\
        \hline
        1 & 83.6 & 91.2 & 91.1 & 83.5 & 90.8 & 90.9 \\
        \rowcolor{Gray} 2 & \textbf{84.0} & 91.6 & \textbf{91.2} & \textbf{83.6} & \textbf{91.3} & 91.0 \\
        4 & \textbf{84.0} & \textbf{91.7} & \textbf{91.2} & 83.4 & 91.1 & \textbf{91.1} \\
        8 & 82.8 & 90.7 & 91.0 & 81.6 & 89.7 & 90.6 \\
        \bottomrule
       \end{tabular}
       \caption{\textbf{Decoder depth}. A shallow decoder with 2 or 4 layers performs better.}
       \label{tab: decoder depth}
    \end{subtable}
    \hfill
    \begin{subtable}{0.49\textwidth}
        \centering
        \begin{tabular}{c|ccc|ccc}
        \toprule
         & \multicolumn{3}{c|}{\textbf{3ML\textsubscript{self}}} & \multicolumn{3}{c}{\textbf{3ML\textsubscript{cross}}} \\
        \textbf{\#dim} &  \textbf{MNLI} & \textbf{QNLI} & \textbf{QQP} & \textbf{MNLI} & \textbf{QNLI} & \textbf{QQP} \\
        \hline
        256 & 83.8 & 91.3 & 91.0 & 83.0 & 90.9 & \textbf{91.1} \\
        \rowcolor{Gray} 512 & 84.0 & \textbf{91.6} & \textbf{91.2} & \textbf{83.6} & \textbf{91.3} & 91.0 \\
        768 & \textbf{84.4} & 91.5 & \textbf{91.2} & 83.3 & 90.7 & 90.9 \\
        1024 & 84.1 & 91.5 & \textbf{91.2} & 83.5 & 90.8 & 91.0 \\
        \bottomrule
       \end{tabular}
        \caption{\textbf{Decoder width}. A decoder with a small hidden dimension still performs well.}
        \label{tab: decoder width}
     \end{subtable}
     \begin{subtable}{0.98\textwidth}
        \centering
        \begin{tabular}{ccc|ccccc|ccccc}
        \toprule
        \multicolumn{3}{c|}{\textbf{Masking Strategy} (\%)} & \multicolumn{5}{c|}{\textbf{3ML\textsubscript{self}}} & \multicolumn{5}{c}{\textbf{3ML\textsubscript{cross}}} \\
        \textbf{masked} & \textbf{replaced} & \textbf{unchanged} & \textbf{MNLI} & \textbf{QNLI} & \textbf{QQP} & \textbf{Avg.} & \textbf{PPL} & \textbf{MNLI} & \textbf{QNLI} & \textbf{QQP} & \textbf{Avg.} & \textbf{PPL} \\
        \hline
        \rowcolor{Gray} 80 & 10 & 10 & 84.0 & \textbf{91.6} & 91.2 & 88.9 & 9.5 & \textbf{83.6} & \textbf{91.3} & \textbf{91.0} & \textbf{88.6} &  9.3\\
        100  & 0 & 0 & 83.1 & 90.7 & 90.9 & 88.2 & 15.6 & 82.4 & 90.3 & \textbf{91.0} & 87.9 & 15.1 \\
        80 & 20 & 0 & 82.5 & 90.4 & 91.0 & 87.9 & 14.0 & 83.1 & 90.7 & 90.9 & 88.2 & \textbf{7.7} \\
        80 & 0 & 20 & \textbf{84.2} & \textbf{91.6} & \textbf{91.3} & \textbf{89.0} & \textbf{7.1} & 82.9 & 90.7 & 90.9 & 88.2 & 14.8 \\
        \bottomrule
       \end{tabular}
        \caption{\textbf{Masking Strategy}. The masking strategies, 80-10-10 and 80-0-20, work better. PPL refers to the validation MLM perplexity.}
        \label{tab: mask rule}
     \end{subtable}
     \caption{\textbf{3ML ablation experiments} on the large model. The default settings are marked in gray, i.e. 2 decoder layers, a hidden dimension of 512 for 3ML's decoder, and a masking rate of $40\%$ with the 80-10-10 strategy. \label{tab: ablation}}
\end{table*}

\section{Efficient Setting}
\label{sec: efficient setting}
Section \ref{sec: prerequisites} shows that two prerequisites for higher efficiency hold for 3MLs. In this section, we further explore different choices of 3ML architecture and the masking rate, trying to select an optimal setting with the trade-off between performance and efficiency.

\subsection{Decoder Architecture}
\label{sec: decoder architecture}
By default, we set the 3ML's encoder with the same hyper-parameter setting as Google BERT. We leave the exploration of the encoder architecture to future work. Since both [MASK]s and unmasked tokens are fed into the 3ML's decoder, it's necessary to select a small decoder for high efficiency.

We explore 3ML's decoders with different numbers of layers and hidden dimensions in Tables \ref{tab: decoder depth} and \ref{tab: decoder width}. As shown in Table \ref{tab: decoder depth}, both 3ML\textsubscript{self} and 3ML\textsubscript{cross} have a similar but surprising finding: A large decoder with 8 layers works the worst, while a small decoder with 2 or 4 layers works the best. We argue that 3ML with a deeper decoder doesn't work well because of our fine-tuning setting. Recapping that 3ML's decoder is removed for fine-tuning, throwing a deeper decoder away means that more pre-trained parameters are removed. For the following experiments, 3ML with a two-layer decoder is the default setting.

Table \ref{tab: decoder width} shows 3ML's performance with different hidden dimensions. 3ML is less sensitive to the hidden dimension of the decoder: the performance of all settings is very similar. By default, we choose a decoder with half of the encoder's dimension for the following experiments.

Both Tables \ref{tab: decoder depth} and \ref{tab: decoder width} show that a small 3ML decoder is enough. It suggests that fusing all information flows after the encoder is easy, again empirically testing our information flow assumption.  

\subsection{Masking Strategy}
\label{sec: masking strategy}
The default masking strategy of vanilla MLM is 80-10-10. I.e. $80\%$ of the corrupted tokens are replaced by [MASK]s, $10\%$ are replaced by random tokens and $10\%$ are kept unchanged. Ideally, we can achieve higher efficiency with the 100-0-0 setting, since we can further reduce the sequence length for 3ML's encoder given the same masking rate as 80-10-10.

We repeat the experiments on masking strategy as BERT in Table \ref{tab: mask rule}, checking whether we have different findings for our 3ML. The default masking strategy 80-10-10 for both 3MLs works much better than 100-0-0. It is also the best strategy for 3ML\textsubscript{cross}. 80-0-20 works similarly to 80-10-10 for 3ML\textsubscript{self}. This finding suggests that keeping prediction on some original (unchanged) tokens is necessary, decreasing the gap between pre-training and fine-tuning. The original BERT paper \cite{DBLP:conf/naacl/DevlinCLT19} had a similar finding. By default, we apply the 80-10-10 masking strategy.

We can further observe the mismatch between pre-training and fine-tuning with the MLM perplexity scores. The prediction of [MASK]s is the hardest for both 3MLs, with the highest perplexity. The prediction of unchanged tokens is the easiest for 3ML\textsubscript{self}. Surprisingly, the prediction of randomly replaced tokens is the easiest for 3ML\textsubscript{cross}, which is contradictory to our intuition that unchanged tokens should be easiest to predict.

\subsection{Masking Rate}
\label{sec: masking rate}
\begin{figure}
  \centering
    \includegraphics[width=0.49\textwidth]{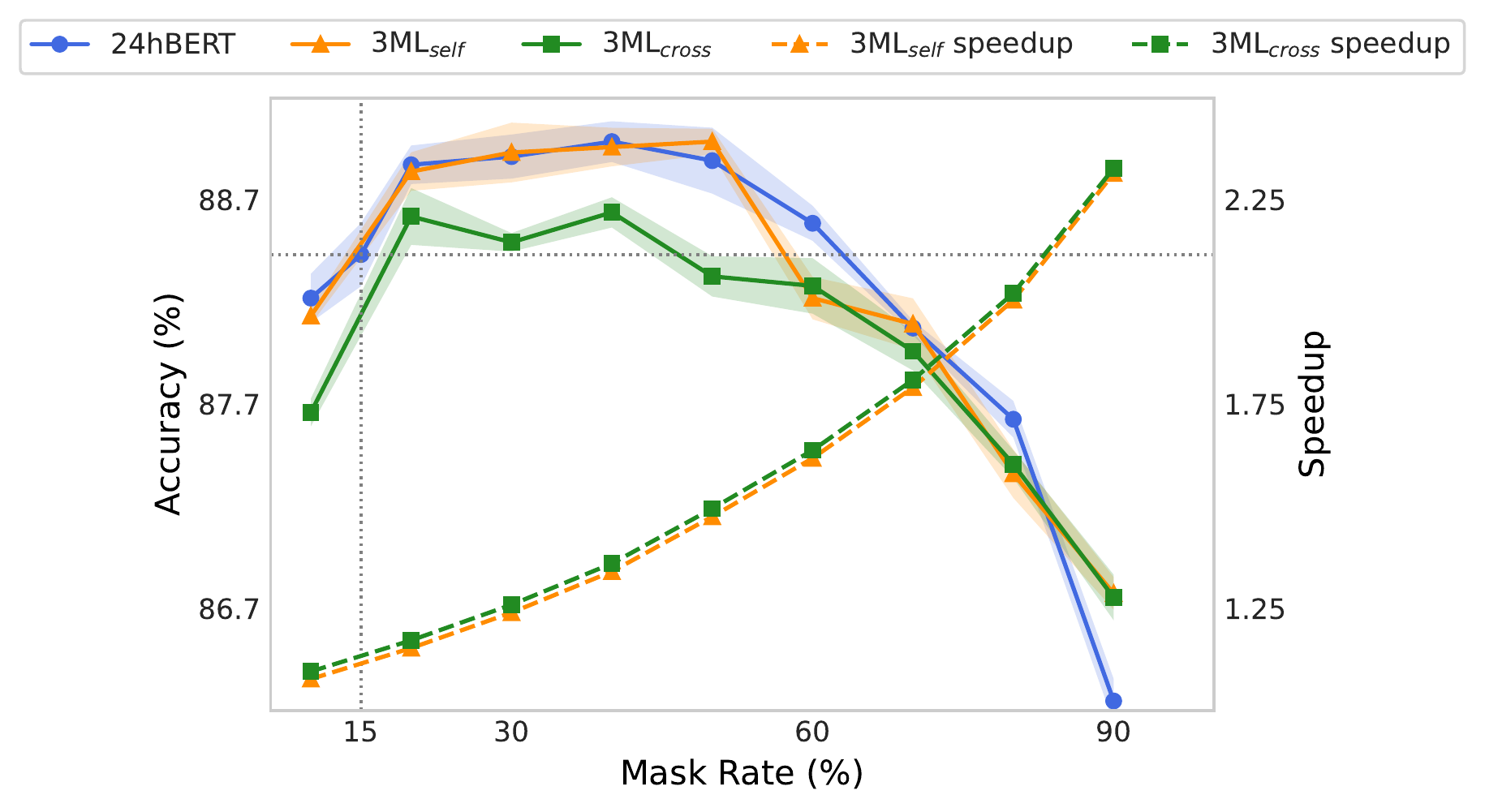}
    \caption{\textbf{Efficiency \textit{vs.} performance}. The best results for 3ML\textsubscript{self} and 3ML\textsubscript{cross} are obtained with a masking rate of $50\%$ and $40\%$, respectively. The accuracy is the macro average accuracy for MNLI, QNLI, and QQP. Speedup is calculated based on the pre-training FLOPs with 24hBERT as the baseline. Experiments are conducted on the large model.}
    \label{fig: efficiency}
\end{figure}

A large masking rate is a necessary prerequisite for the efficiency of 3ML. However, a too-large masking rate hurts the performance as shown in Figure \ref{fig: 24hbert mask more}. In this section, we study 3ML's trade-off between efficiency and performance. From Figure \ref{fig: efficiency}, we achieve higher speedups with increasing masking rates. With a masking rate of 50\%, we can obtain near 1.5 times speedup, saving 1/3 of the computation budget.

Both 3ML\textsubscript{self} and 3ML\textsubscript{cross} share a similar trend as 24hBERT: better accuracy is obtained with a middle-level masking rate. 3ML\textsubscript{self} works almost the same as 24hBERT with a masking rate of $<40\%$. A masking rate of $50\%$ works best for 3ML\textsubscript{self}, even better than the best 24hBERT. It means that we can save 1/3 of the original training budget without any performance drop and even with slight improvement. 3ML\textsubscript{cross} works slightly worse than 24hBERT with the same masking rate. It performs best with a masking rate of $40\%$ and better than 24hBERT with the default masking rate ($15\%$). The training FLOPs of 24hBERT are 1.37 times larger than the ones of 3ML\textsubscript{cross} when the masking rate is $40\%$. We recommend choosing a masking rate of 40\% or 50\% for both 3MLs for obtaining good performance and high efficiency.

\begin{table*}[h]
\scriptsize
\centering
\begin{tabular}{lccccccccccc}
\toprule
\textbf{Model} & \textbf{\#Params} & \textbf{Speedup}  & \textbf{CoLA} & \textbf{SST} &  \textbf{MRPC} & \textbf{STS} & \textbf{QQP} & \textbf{MNLI-m/mm} & \textbf{QNLI} & \textbf{RTE} & \textbf{Avg.}\\
& M &  & Mcc & Acc & F1/Acc & Pear/Spea & F1/Acc & Acc & Acc & Acc \\
\hline
\multicolumn{12}{l}{\textit{Efficient pre-training recipe, large model}} \\
24hBERT-15\%$^\ast$ & 355 & 1.00$\times$  & 61.3 & 93.0 & 92.0/89.0 & 89.4/89.2 & 88.0/91.1 & 84.0/83.6 & 90.4 & 76.9 & 84.3 \\
24hBERT-40\%$^\ast$ & 355 & 1.00$\times$  & 61.4 & \textbf{93.2} & \textbf{92.1/89.2} & \textbf{89.5/89.5} & 88.0/91.2 & \textbf{84.4/84.3} & 91.3 & 80.1 & \textbf{85.0} \\
3ML\textsubscript{self}-40\% & 362 & 1.34$\times$ & 61.5 & 92.3 & 92.1/89.0 & \textbf{89.5/89.5} & \textbf{88.1/91.2} & \textbf{84.4/84.3} & \textbf{91.6} & \textbf{80.9} & \textbf{85.0}  \\
3ML\textsubscript{self}-50\% & 362 & 1.47$\times$ & \textbf{62.4} & 92.3 & 92.0/89.2 & \textbf{89.5/89.5} & \textbf{88.1/91.2} & 84.1/84.4 & 91.4 & 79.8 & \textbf{85.0} \\
3ML\textsubscript{cross}-40\% & 364 & 1.37$\times$ & \textbf{62.4} & 92.8 &  91.2/87.5 & 89.0/88.8 & 87.9/91.0 & 83.7/83.5 & 91.3 & 80.5 & 84.8 \\
\hline
\multicolumn{12}{l}{\textit{Longer pre-training recipe, base model}} \\
BERT & 110 & 1.15$\times$ & 54.3 & 91.5 & 89.5 & 88.9 & 89.8 & 83.5 & 91.2 & 71.1 & 82.5  \\
RoBERTa$^\ast$ & 125 & 1.00$\times$ & 61.3 & \textbf{93.5} & 91.5/88.8 & 89.7/89.5 & \textbf{88.1/91.3} & 84.7/84.9 & 91.4 & 78.9 & 84.9  \\
ELECTRA & 126 & 1.15$\times$ & - & - & -/-$^{\circ}$ &  -/-$^{\circ}$ & -/-$^{\circ}$ & - & - & - & 85.1 \\
3ML\textsubscript{self}-40\%-125K$^{\diamond}$ & 129 & 1.22$\times$ & 60.6 & 92.9 & \textbf{93.2/90.7} & 89.9/89.7 & 87.7/90.9 & 84.5/84.3 & 91.3 & 78.7 & 84.9 \\
3ML\textsubscript{self}-40\%-153K$^{\diamond}$ & 129 & 1.00$\times$ & 61.4 & 92.9 & 92.3/89.2 & \textbf{90.3/90.1} & 87.9/91.1 & \textbf{84.8/84.9} & 91.4 & \textbf{81.2} & \textbf{85.3} \\
3ML\textsubscript{self}-50\%-125K$^{\diamond}$ & 129 & 1.28$\times$ & \textbf{63.1} & 92.4 & 92.0/89.0 & 89.7/89.4 & 87.6/90.9 & 84.1/84.3 & 91.9 & 79.1 & 85.0 \\
3ML\textsubscript{self}-50\%-160K$^{\diamond}$ & 129 & 1.00$\times$ & 61.7 & 93.3 & 92.9/90.2 & 90.1/89.9 & 87.9/91.1 & 84.3/84.5 & 91.3 & 79.1 & 85.1 \\
3ML\textsubscript{cross}-40\%-157K$^{\diamond}$ & 130 & 1.00$\times$ & 56.2 & 92.3 & 91.8/88.7 & 89.6/89.3 & 87.6/90.8 & 82.7/83.6 & 90.8 & 78.3 & 83.7 \\

\bottomrule
\end{tabular}
\caption{\label{tab: full glue dev}
\textbf{Complete results on GLUE dev. set}. Speedup is computed based on the pre-training FLOPs. ``$^\ast$'' denotes our re-implementation.  ``$^{\circ}$'' means the corresponding metric is used for calculating the average score. When using the same metrics as ELECTRA to compute the average score, models with ``$^{\diamond}$'' stay the same as the shown average score. More details on the baseline models are in Appendix \ref{sec: differences among baselines}.}
\end{table*}

\section{Comparison with Previous Work}
\label{sec: compare with previous work}
In this section, we compare 3ML with strong baselines on the development set of GLUE. We list the results of both efficient and longer pre-training recipes in Table \ref{tab: full glue dev}.

\subsection{Result of Efficient Pre-training Recipe}
\label{sec: efficient pre-training recipe}
The first block of Table \ref{tab: full glue dev} shows results from the efficient pre-training recipe. We train all models with the same number of updates and an identical learning rate. So the improvement shown here is not obtained by seeing more data or doing more extensive gradient updates.  

Similar to Figure \ref{fig: 24hbert mask more}, 24hBERT with a masking rate of 40\% performs consistently better than the one with 15\% on all GLUE tasks, achieving 0.7\% absolute improvement on average. It further suggests that masking more is possible and favorable. 3ML\textsubscript{self} with a masking rate of 40\% and 50\% has the same average score as 24hBERT-40\%. But it requires less computation, to be precise only 75\% and 68\% of the original computation budget. 3ML\textsubscript{cross} achieves a slightly worse result than 24hBERT-40\%, losing 0.2\% performance on average, but being slightly more efficient than 3ML\textsubscript{self} with the same masking rate.

Compared to the number of trainable parameters of 24hBERT, the increasing parameters for 3ML due to an extra decoder are negligible, only accounting for 2\% of all 24hBERT parameters. In addition, 3ML's decoder is discarded for fine-tuning. We believe that the well-performed 3MLs benefit from our two-stage training method rather than the slightly added parameters.

\subsection{Result of Longer Pre-training Recipe}
\label{sec: longer pre-training recipe}
3ML behaves well and efficiently on a limited computation budget. We are also interested in its scaling behavior which is critical for a language model. As we normally train a language model on large data for better generalization on a wide range of tasks. Due to our limited computation resources, we only implement the scaling experiment on the base model with more updates till convergence. We leave the scaling experiments on a bigger data set and a larger model to future work.

The results from the longer pre-training recipe are shown in the second block of Table \ref{tab: full glue dev}. When seeing the same amount of data (125K updates), 3ML\textsubscript{self} performs better than BERT (84.9/85.0 vs. 82.5), comparable to RoBERTa (84.9/85.0 vs. 84.9) and ELECTRA (84.9/85.0 vs. 85.1). However, 3ML-125K is faster than any baseline (28\% faster than RoBERTa with a masking rate of 50\%). When using the same computation budget as RoBERTa, 3ML\textsubscript{self}-$40\%$-153K achieves the best result (85.3) among all models. We also notice that the pre-training of 3ML\textsubscript{self}s already converge with 125K updates. As also shown in Table \ref{tab: full glue dev}, we only obtain 0.4\% and 0.5\% improvement with extra 28K and 35K updates for 3ML\textsubscript{self}-40\% and 3ML\textsubscript{self}-50\%, respectively. Further improvement is expected if the model is trained on a larger dataset. 

Similar to the efficient pre-training recipe, 3ML\textsubscript{cross} performs worse than the strong baselines, RoBERTa and ELECTRA, but better than BERT-base. We argue the reason for the poor performance of 3ML\textsubscript{cross} is: The positional embeddings for [MASK]s and unmasked tokens are not in the same latent space (Figure \ref{fig: 3ml architecture}), which makes it difficult for the model to fuse all information and predict missing information. In addition, the decoder doesn't further transfer information among unmasked tokens (the most complex flow). We leave the further investigation to future work.  

In summary, when trained on the same number of samples, 3ML\textsubscript{self} performs similarly to strong baselines with less computation. 3ML\textsubscript{cross} performs comparably to the baselines for efficient pre-training. When trained with a similar amount of computation, 3ML\textsubscript{self} performs the best and 3ML\textsubscript{cross} performs better than the standard baseline, BERT. 

\section{Related Work}
The most related work to this paper is MAE \cite{DBLP:journals/corr/abs-2111-06377} from computer vision. 3ML\textsubscript{self} shares almost the same architecture as MAE, but with the additional prediction of unchanged tokens and randomly replaced tokens, while MAE only reconstructs the masked patches. We also have different findings from MAE: A small decoder works better for 3ML\textsubscript{self}. While MAE applied a deeper decoder for better performance.

3ML's encoder-decoder architecture looks similar to BART \cite{DBLP:conf/acl/LewisLGGMLSZ20} and MASS \cite{DBLP:conf/icml/SongTQLL19}. But BART and MASS apply a causal masking decoder that is suitable for generation tasks rather than classification tasks. Our decoder is still a bidirectional architecture like the encoder. In addition, the decoder of BART and MASS is used for fine-tuning downstream tasks. For GLUE tasks, one needs to input the same sequence to both encoder and decoder, which is less efficient. 3ML's decoder is removed for fine-tuning, having the same inference speed as vanilla MLM.

\citet{hou-etal-2022-token} do a concurrent work, dropping the representations of unimportant tokens for some intermediate layers to reduce the sequence length for efficiency. We don't include any prior information and only drop the [MASK] token at the very beginning. In addition, 3ML achieves better results.

\citet{DBLP:journals/corr/abs-2202-08005} and our work have the same finding, better performance for a middle-level masking rate. Although ALBERT \cite{DBLP:conf/iclr/LanCGGSS20} and ELECTRA \cite{DBLP:conf/iclr/ClarkLLM20} make the pre-training of MLM more efficient, their studies are orthogonal to ours.

\section{Conclusion}
\label{sec: conclusion}
We propose a two-stage learning method for efficient masked language modeling and design two models, 3ML\textsubscript{self} and 3ML\textsubscript{cross}, for our method. Two prerequisites for our efficient method are: We can have a higher masking rate and append [MASK]s at a later layer. Our experiments show that both, masking more and masking later, are possible and favorable. This allows us to reduce the sequence length of the encoder during pre-training by a factor depending on the masking rate. By conducting extensive experiments, we observe that 3ML\textsubscript{self} performs better on downstream tasks than 3ML\textsubscript{cross}. It can speed up the pre-training by a factor of 1.5x for our efficient pre-training recipe without any performance degradation. Using roughly the same computation budget, 3ML\textsubscript{self} outperforms all of our strong baselines like ELECTRA and RoBERTa.

\section*{Limitations}
Our investigation is limited to classification tasks. While 3ML outperforms other models on GLUE tasks, it might not be good at other tasks, especially for the mask-infilling task where the token embedding of [MASK] is used directly. More tasks need to be evaluated to validate 3ML's robustness. 

In addition, we only train 3MLs on BookCorpus and English Wikipedia. The scaling behavior of 3ML with respect to model size and amount of data is an open question. Further, it needs to be validated whether the results transfer to other languages than English. We leave this to the future. We also don't apply any fine-tuning tricks, like layer-wise learning rate in ELECTRA, and tune the hyperparameters. It would be better for us to provide a specific optimal recipe for our model, making it more practical. 

\section*{Acknowledgments}
The authors from RWTH Aachen University partially received funding from eBay Inc. We thank the colleagues from eBay (Michael Kozielski and Shahram Khadivi) and RWTH Aachen University (Yingbo Gao and Christian Herold) for their discussion.

\bibliography{anthology,custom}
\bibliographystyle{acl_natbib}

\appendix

\section{Pre-training Hyperparameters}
\label{sec: pretraining hyperparameters}
Pre-training hyperparameters are shown in Table \ref{tab: pretraining hyperparameters}. Settings for large and base models are mainly borrowed from 24hBERT \cite{izsak-etal-2021-train} and RoBERTa \cite{DBLP:journals/corr/abs-1907-11692}, respectively. 

\begin{table*}
  \centering
    \begin{tabular}{lcc}
    \toprule
    \textbf{Hyperparameter} & \textbf{3ML-large} & \textbf{3ML-base} \\
    \hline
    \textit{Encoder} \\
    Number of Layers & 24 & 12 \\
    Hidden Size & 1024 & 768 \\
    FFN Inner Hidden Size & 4096 & 3072 \\
    Attention Heads & 16 & 12 \\
    Attention Head Size & 64 & 64 \\
    \hline 
    \textit{Decoder} \\
    Number of Layers & 2 & 2 \\
    Hidden Size & 512 & 384 \\
    FFN Inner Hidden Size & 2048 & 1536  \\
    Attention Heads & 8 & 6 \\
    Attention Head Size & 64 & 64 \\
    \hline
    \textit{Whole model} \\
    Dropout & 0.1 & 0.1 \\
    Attention Dropout & 0.1 & 0.1 \\
    Layer Normalization & pre-LN & post-LN 3ML\textsubscript{self} / pre-LN 3ML\textsubscript{cross} \\
    Sequence Length & 128 & 512 \\
    \hline
    \textit{Optimizer} \\
    Warmup Proportion & 0.06 & 0.06 \\
    Peak Learning Rate & 2e-3 & 7e-4 \\
    Batch Size & 4096 & 2048 \\
    Weight Decay & 0.01 & 0.01 \\
    Max Steps & 23K & 153K for 3ML\textsubscript{self} / 160K 3ML\textsubscript{cross} \\
    Learning Rate Decay & Linear & Linear \\
    Adam $\epsilon$ & 1e-6 & 1e-6 \\
    Adam $(\beta_1, \beta_2)$ & (0.9, 0.98) & (0.9, 0.98) \\
    Gradient Clipping & 0.0 & 0.0 \\
    \bottomrule
\end{tabular}
\caption{\label{tab: pretraining hyperparameters}
\textbf{Pre-training hyperparameters} for both 3ML\textsubscript{self} and 3ML\textsubscript{cross}. Pre-LN 3ML\textsubscript{cross} is more stable than Post-LN. If the gradient explodes, set gradient clipping as 1.0 instead.}
\end{table*}

\section{Fine-tuning Hyperparameters}
\label{sec: finetuning hyperparameters}
Fine-tuning hyperparameters for GLUE are shown in Table \ref{tab: finetuning hyperparameters}, borrowed from 24hBERT \cite{izsak-etal-2021-train}. For each task, we run the fine-tuning on the whole search space five times with different random seeds, then select the best values from each running and finally choose the median of these five values.

\begin{table*}
  \centering
    \begin{tabular}{lcc}
    \toprule
    \textbf{Hyperparameter} & \textbf{QQP, MNLI, QNLI} & \textbf{CoLA, SST, MRPC, STS, RTE} \\
    \hline
    Learning Rate & \{5e-5, 8e-5\} & \{1e-5, 3e-5, 5e-5, 8e-5\} \\
    Batch Size & 32 & \{16, 32\} \\
    Weight Decay & 0.1 & 0.1 \\
    Max Epochs & \{3, 5\} & \{3, 5, 10\} \\
    Warmup Proportion & 0.06 & 0.06 \\
    \bottomrule
\end{tabular}
\caption{\label{tab: finetuning hyperparameters}
\textbf{Fine-tuning hyperparameters} for both 3ML\textsubscript{self} and 3ML\textsubscript{cross}. Same as RoBERTa \cite{DBLP:journals/corr/abs-1907-11692}, RTE, STS, and MRPC are fine-tuned from the MNLI model instead of the pre-trained model.
}
\end{table*}

\section{Differences among Baselines}
\label{sec: differences among baselines}
We use some base models (BERT, RoBERTa, and ELECTRA) as our baselines for the longer pre-training recipe. We specify the main differences among them here. RoBERTa contains 15M more trainable parameters (see Table \ref{tab: full glue dev}) than BERT because they apply different sub-word algorithms and have different vocabulary sizes, 50K for RoBERTa, and 30K for BERT (also for ELECTRA). The extra 16M parameters (compared to BERT) from ELECTRA come from the generator, while its discriminator shares the same architecture as BERT. In addition, BERT, ELECTRA, and RoBERTa are trained with a batch size of 256, 256, and 2048, respectively. Their number of updates is 1M, 766K, and 125K, respectively. In another word, BERT and RoBERTa see the same amount of data (256M samples), while ELECTRA sees less data (197M samples). But RoBERTa conducts the least updates. Among these three models, the training recipe from RoBERTa is more suitable for scaling. We can train an MLM quickly with a larger batch size by allocating a large number of GPUs. Therefore, we borrow the training recipe from RoBERTa as our longer pre-training recipe.

\section{Calculation of FLOPs}
\label{sec: FLOPs}
We borrow a simplified version of FLOPs calculation from \citet{DBLP:conf/iccv/PanZLH021} where the computation for bias, activation, and dropout is neglected because it only occupies a small amount ($<1\%$) of the total FLOPs. We restate their calculation and make some modifications here. The meaning for different notations is listed in Table \ref{tab: notation} for your convenience.

\textbf{Transformer block with self-attention} Given the sequence length $n$ and the embedding dimension $d$, the FLOPs of the multi-head self-attention (MSA) layer come from: (1) the projection of the input sequence to key, query and, value $\phi_{qkv} = 2 \cdot 3nd^2$ \footnote{We double all original calculation by considering the fused multiply-add ops \cite{DBLP:conf/iclr/ClarkLLM20}.}; (2) the attention map from key and query $\phi_{map} = 2n^2d$; (3) the self-attention operation $\phi_{attn} = 2n^2d$; (4) the projection of the self-attention output $\phi_{out} = 2n^2d$. Then the overall FLOPs for an MSA layer are:
\begin{align*}
    \phi_{MSA}(n, d) = 8nd^2 + 4n^2d
\end{align*} 
There are two fully-connected (FC) layers for a feed-forward (MLP) layer. The first FC projects the embedding dimension from $d$ to $4d$, and the second one project it back to $d$. The FLOPs are:
\begin{align*}
    \phi_{MLP}(n, d) = 2 \cdot 8nd^2 = 16nd^2
\end{align*}
So the overall FLOPs for a transformer block are:
\begin{align*}
    \phi_{BLK}(n, d) &= \phi_{MSA}(n, d) + \phi_{MLP}(n, d)  \\
    &= 24nd^2 + 4n^2d
\end{align*}

\textbf{Embedding layer} We assume all models, including baseline models, implement sparse lookup for token embedding, which is different from ELECTRA \cite{DBLP:conf/iclr/ClarkLLM20} which states RoBERTa and BERT obtain the token embedding by multiplying the embedding layer with one-hot vectors. We make this assumption since any model can be easily re-implemented in this efficient way. Sparse lookup is efficient. It can be neglected for calculating FLOPs.

\textbf{Prediction layer} The prediction layer consists of two FC layers. The first FC layer keeps the input and output sequence in the same dimension, and the second one projects to the vocabulary size $|V|$. Like typical language models, we use shared weight between the embedding layer and the second FC layer. Like RoBERTa, we only input the masked tokens, including unchanged and randomly replaced tokens, to the prediction layer. With a masking rate of $r$, the FLOPs are:
\begin{align*}
    \phi_{Pred}(n, r, d, |V|) = 2nr(d^2 + d|V|)
\end{align*}

\textbf{RoBERTa} The total FLOPs for RoBERTa can be computed as
\begin{align}
\label{FLOPs of RoBERTa}
    &\phi_{RoBERTa}(b, u, l, n, d, r, |V|) \nonumber \\
    =& 2bu(l \cdot \phi_{BLK}(n, d) + \phi_{Pred}(n, r, d, |V|))
\end{align}
where $b$ is the batch size, $u$ is the number of updates, $l$ is the number of transformer blocks and $2$ at the beginning denotes the forward and backward process. Both forward and backward processes consume similar FLOPs. 

\textbf{3ML\textsubscript{self}} has two types of transformer blocks, one for the encoder and one for the decoder. The only difference between them is the hidden dimension. To project the output sequence from the encoder to the same hidden dimension of the decoder, we add an FC layer in between. The FLOPs of the prediction layer are modified as:
\begin{align*}
     & \phi_{Pred}^{'}(n, r, d_{en}, d_{de}, |V|) \\
    = & 2nr(d_{de}d_{en} + d_{en}|V|)
\end{align*}
where $d_{en}$ and $d_{de}$ is the hidden dimension of the encoder and decoder, respectively. Then the overall FLOPs are:
\begin{align}
\label{FLOPs for 3ML_self}
    &\phi_{3ML\textsubscript{self}}(b, u, l_{en}, l_{de}, n_{en}, n, d_{en}, d_{de}, r, |V|) \nonumber \\
    =& 2bu(2n_{en}d_{en}d_{de} + l_{en} \cdot \phi_{BLK}(n_{en}, d_{en}) \nonumber \\
    & + l_{de} \cdot \phi_{BLK}(n, d_{de}) \nonumber \\
    & + \phi_{Pred}^{'}(n, r, d_{en}, d_{de}, |V|))
\end{align}
where $n_{en} = (1-0.8r)n$ since only 80\% of all masked tokens are replaced by [MASK]s. The first term on the right is for the dimension projection from the encoder to the decoder. $l_{en}$ and $l_{de}$ are the number of Transformer blocks for the encoder and decoder, respectively.

\textbf{Transformer block with self \& cross-attention} The decoder of 3ML\textsubscript{cross} contains an MSA layer, a multi-head cross-attention (MCA) layer, and an MLP layer. The FLOPs calculation of both MSA and MLP stay the same as above. Similar to the MSA layer, the FLOPs for different components of the MCA layer are
\begin{align*}
    \phi_{qkv}^{'} &= 4n_{en}d_{de}^{2} + 2n_{de}d_{de}^{2} \\
    \phi_{map}^{'} &= 2n_{en}n_{de}d_{de} \\
    \phi_{attn}^{'} &= 2n_{en}n_{de}d_{de} \\
    \phi_{out}^{'} &= 2n_{de}d_{de}^2
\end{align*}
where $n_{en}$ and $n_{de}$ are the sequence lengths of the encoder and decoder inputs, respectively. So the FLOPs for an MCA layer are:
\begin{align*}
    & \phi_{MCA}(n_{en}, n_{de}, d_{de}) \\
    =& 4n_{en}d_{de}^{2} + 4n_{de}d_{de}^{2} + 4n_{en}n_{de}d_{de}
\end{align*}
Then the overall FLOPs for a cross-attention Transformer block are:
\begin{align*}
    & \phi_{CBLK}(n_{en}, n_{de}, d_{de}) \\
    =& \phi_{BLK}(n_{de}, d_{de}) + \phi_{MCA}(n_{en}, n_{de}, d_{de}) \\
    =& 4n_{en}d_{de}^{2} + 28n_{de}d_{de}^{2} + 4n_{en}n_{de}d_{de} + 4n_{de}^2d_{de}
\end{align*}

\textbf{3ML\textsubscript{cross}} Similar to 3ML\textsubscript{self}, 3ML\textsubscript{cross} has an encoder and a decoder with a hidden size of $d_{en}$ and $d_{de}$, respectively. Since their hidden sizes might be different, there is an FC layer projecting from $d_{en}$ to $d_{de}$. Then the overall FLOPs for training a 3ML\textsubscript{cross} is
\begin{align}
\label{FLOPs for 3ML_cross}
    &\phi_{3ML\textsubscript{cross}}(b, u, l_{en}, l_{de}, n_{en}, n_{de}, d_{en}, d_{de}, |V|) \nonumber \\
    =& 2bu(2n_{en}d_{en}d_{de} + l_{en} \cdot \phi_{BLK}(n_{en}, d_{en}) \nonumber \\
    & + l_{de} \cdot \phi_{CBLK}(n_{en}, n_{de}, d_{de}) \nonumber \\
    & + \phi_{Pred}(n_{en}, \frac{0.2r}{1-0.8r}, d_{en}, |V|)) \nonumber \\
    & + \phi_{Pred}^{'}(n_{de}, 1, d_{en}, d_{de}, |V|))
\end{align}
where $n_{de} = 0.8nr$ and $n_{de} = (1- 0.8r)n$, since only 80\% of all masked tokens are replaced by [MASK]s. The second last term denotes the prediction layer on the encoder side. Both prediction layers share the weight from the embedding layer. The masking rate from the second last term is achieved by calculating the ratio between the number of unchanged and randomly replaced tokens and $n_{en}$. The masking rate of the last term is $1$ because there are only [MASK]s on the decoder side.

\begin{table}
  \centering
    \begin{tabular}{ll}
    \toprule
    \textbf{Notation} & \textbf{Explanatiorebib} \\
    \hline
    $n$ & sequence length \\
    $n_{en}$ & sequence length for encoder \\ 
    $n_{de}$ & sequence length for decoder \\
    $d$ & hidden dimension \\
    $d_{en}$ & hidden dimension for encoder \\
    $d_{de}$ & hidden dimension for decoder \\
    $r$ & masking rate \\
    $|V|$ & vocabulary size \\
    $b$ & batch size \\
    $u$ & \#updates \\
    $l$ & \#(Transformer block) \\
    $l_{en}$ & \#(Transformer block) for encoder \\
    $l_{en}$ & \#(Transformer block) for decoder \\
    \bottomrule
\end{tabular}
\caption{\label{tab: notation} Notation for the calculation of training FLOPs.}
\end{table}

\end{document}